\pdfoutput=1
\documentclass[11pt,a4paper]{article}
\usepackage[hyperref]{acl2021}
\usepackage{times}
\usepackage{latexsym}
\usepackage{url}
\usepackage{amsmath}
\usepackage{multirow}
\usepackage{xcolor}

\newcommand\numberthis{\addtocounter{equation}{1}\tag{\theequation}}

\definecolor{ForestGreen}{RGB}{34,139,34}
\aclfinalcopy 
\def\aclpaperid{3751} 

\title{XL-Sum: Large-Scale Multilingual Abstractive Summarization for 44 Languages}

\author{
Tahmid Hasan$^1$\thanks{~ These authors contributed equally to this work.} , Abhik Bhattacharjee$^1$\footnotemark[1] ,  Md Saiful Islam$^2$, Kazi Samin$^1$,\\
\textbf{Yuan-Fang Li}$^3$, \textbf{Yong-Bin Kang}$^4$, \textbf{M. Sohel Rahman}$^1$, \textbf{Rifat Shahriyar}$^1$\\ [3pt]
Bangladesh University of Engineering and Technology (BUET)$^1$, University of Rochester$^2$,\\
Monash University$^3$, Swinburne University of Technology$^4$\\ [3pt]
\texttt{tahmidhasan@cse.buet.ac.bd}, \texttt{\{abhik,samin\}@ra.cse.buet.ac.bd},\\
\texttt{mislam6@ur.rochester.edu}, \texttt{yuanfang.li@monash.edu},\\
\texttt{ykang@swin.edu.au}, \texttt{\{msrahman,rifat\}@cse.buet.ac.bd}
}

\date{}

\begin{document}

\maketitle

\begin{abstract}

Contemporary works on abstractive text summarization have focused primarily on high-resource languages like English, mostly due to the limited availability of datasets for low/mid-resource ones. In this work, we present XL-Sum, a comprehensive and diverse dataset comprising 1 million professionally annotated article-summary pairs from BBC, extracted using a set of carefully designed heuristics. The dataset covers 44 languages ranging from low to high-resource, for many of which no public dataset is currently available.  XL-Sum is highly abstractive, concise, and of high quality, as indicated by human and intrinsic evaluation. We fine-tune mT5, a state-of-the-art pretrained multilingual model, with XL-Sum and experiment on multilingual and low-resource summarization tasks. XL-Sum induces competitive results compared to the ones obtained using similar monolingual datasets: we show higher than 11 ROUGE-2 scores on 10 languages we benchmark on, with some of them exceeding 15, as obtained by multilingual training. Additionally, training on low-resource languages individually also provides competitive performance. To the best of our knowledge, XL-Sum is the largest abstractive summarization dataset in terms of the number of samples collected from a single source and the number of languages covered. We are releasing our dataset and models to encourage future research on multilingual abstractive summarization. The resources can be found at \url{https://github.com/csebuetnlp/xl-sum}.

\end{abstract}

\begin{table}[h]
\centering
\begin{tabular}{p{0.98\linewidth}}
\hline
\hline
\textbf{Input Article:} \textcolor{red}{Yahoo's patents suggest} users could weigh the type of ads against the sizes of discount before purchase. It \textcolor{violet}{says in two US patent applications} that ads for digital book readers have been ``less than optimal" to date. [...] ``Greater levels of advertising, which may be more valuable to an advertiser and potentially more distracting to an e-book reader, may warrant higher discounts," it states. [...] It adds that the more willing the customer is to see the ads, the \textcolor{brown}{greater the potential} discount. [...] At present, several Amazon and Kobo \textcolor{ForestGreen}{e-book readers offer full-screen adverts} when the device is switched off and show smaller ads on their menu screens. [...] Yahoo does not currently provide ads to these devices, and a move into the area could \textcolor{blue}{boost its shrinking revenues}.\\

\hline
\hline
\textbf{Summary:} \textcolor{red}{Yahoo has \underline{signalled}} \textcolor{violet}{it is \underline{investigating}} {\color{ForestGreen} e-book adverts} \textcolor{brown}{\underline{as a way} to} \textcolor{blue}{\underline{stimulate} its \underline{earnings}}. \\
\hline
\hline
\end{tabular}
\caption{A sample article-summary pair from the XL-Sum dataset. To highlight the abstractiveness of the summary, we underline the novel words and phrases that do not appear in the article text. Also, portions of the article relevant to the summary have been color-coded. As we can see, these portions are concisely paraphrased in the summary, unlike extractive methods.
\iffalse \yb{Move Table 1 to the page being discussed.}\fi \iffalse\msi{Alternatively, add a self-explanatory caption with relevant information, put the novel words/phrases in a different color/highlight, to establish that this is an abstractive summary.}\fi}
\label{tab:example}
\vspace{-5mm}
\end{table}

\section{Introduction}

Automatic text summarization \citep{nenkova2011automatic} is a fundamental problem in natural language processing. Given an input text (typically a long document or article), the goal is to generate a smaller, concise piece of text that conveys the key information of the input text. There are two main approaches to automatic text summarization: \textit{extractive} and \textit{abstractive}. Extractive methods crop out one or more segments from the input text and concatenate them to produce a summary. These methods were dominant in the early era of summarization, but they suffer from some limitations, including weak coherence between sentences, inability to simplify complex and long sentences, and unintended repetition~\citep{see-etal-2017-get, widyassari2020review}. 

Abstractive summarization, on the other hand, generates summaries that may contain words and phrases not present in the input text (e.g., via paraphrasing), and may arguably relate more to human-generated summaries \citep{hsu2018unified}. 
Although abstractive summaries can be more coherent and concise than extractive summaries \citep{cohn2008sentence}, generating them is more challenging due to the nature of this task. Limited availability of good datasets conducive to abstractive methods has made it even more difficult. For these reasons, extractive models had been performing better than abstractive ones historically. However, the success of sequence-to-sequence (seq2seq) models \citep{cho2014learning, sutskever2014sequence} over the last decade and the recent advances in Transformer-based models \citep{vaswani2017attention, devlin2018bert} have rejuvenated abstractive text summarization \citep{rush2015neural, see-etal-2017-get, zhang2020pegasus}, which had previously received much less attention in comparison to extractive approaches \citep{nenkova2012survey}. Still, the scarcity of good datasets, especially for low-resource languages, remains a roadblock.

\begin{table*}[h]
\begin{minipage}{\textwidth}
\centering
\begin{tabular}{lrlrlr}
\hline
\textbf{Language} & \textbf{\#Samples} & \textbf{Language} & \textbf{\#Samples} & \textbf{Language} & \textbf{\#Samples}\\
\hline
Amharic & 5,461 & Korean & 4,281 & Somali & 5,636\\
Arabic & 40,327 & Kyrgyz & 2,315 & Spanish & 44,413 \\
Azerbaijani & 7,332 & Marathi & 11,164 & Swahili & 10,005 \\
Bengali & 8,226 & Nepali & 5,286 & Tamil & 17,846 \\
Burmese & 5,002 & Oromo & 5,738 & Telugu & 11,308 \\
Chinese & 39,810 & Pashto & 15,274 & Thai & 6,928 \\
English & 301,444 & Persian & 25,783 & Tigrinya & 4,827 \\
French & 9,100 & Pidgin\footnote{West African Pidgin English} & 9,715 & Turkish & 29,510 \\
Gujarati & 9,665 & Portuguese & 23,521 & Ukrainian & 57,952 \\
Hausa & 6,313 & Punjabi & 8,678 & Urdu & 40,714 \\
Hindi & 51,715 & Russian & 52,712 & Uzbek & 4,944 \\
Igbo & 4,559 & Scottish Gaelic & 1,101 & Vietnamese & 23,468 \\
Indonesian & 44,170 & Serbian (Cyrillic) & 7,317 & Welsh & 11,596 \\
Japanese & 7,585 & Serbian (Latin) & 7,263 & Yoruba & 6,316 \\
\cline{5-6}
Kirundi & 5,558 & Sinhala & 3,414 & \textbf{Total} & \textbf{1,005,292} \\
\hline
\end{tabular}
\caption{Languages covered by the XL-Sum dataset, and the number of samples for each language. Here, a sample denotes an article-summary pair. If we consider languages with less than 15,000 training samples to be low-resource, then more than two-thirds of the constituent languages in XL-Sum fall into this category.}\label{tab:dataset}

\end{minipage}
\end{table*}

Typical seq2seq models are heavily data-driven, i.e., a large number of article-summary pairs are required to train them effectively. As a result, abstractive summarization has centered around the English language, as most large abstractive summarization datasets \citep{hermann2015teaching, grusky2018newsroom, narayan2018don} are available in English only. Though there have been some recent efforts for curating multilingual abstractive summarization datasets \citep{giannakopoulos2015multiling, cao2020multisumm, scialom2020mlsum}, they are limited in terms of the number of languages covered, the number of training samples, or both.

In this work, we introduce \textbf{XL-Sum}, a large-scale abstractive summarization dataset of news articles crawled from the British Broadcasting Corporation (BBC)\footnote{\url{https://www.bbc.com/}} website. Using a custom crawler, we collect 1 million professionally annotated article-summary pairs covering 44 languages. Originating from a single source, these samples exhibit similar summarization strategies across all languages, making them ideal for the multilingual summarization task. XL-Sum introduces the first publicly available summarization dataset and benchmarks for many languages (e.g., Bengali, Swahili). Thus, this dataset potentially enables and facilitates research on low-resource languages, bringing technological advances to communities of these languages that have been traditionally under-served.

We achieve higher than 11 ROUGE-2 score on the 10 languages we benchmark on multilingual summarization, even exceeding 15 ROUGE-2 score (16.58 being the state-of-the-art for English, obtained by \citet{zhang2020pegasus} on XSum~\citep{narayan2018don}, a similar dataset) on many of them. In addition, we also conduct experiments on low-resource summarization task and show competitive results, indicating that the dataset can be used even for the low-resource languages individually.


In summary, we make the following main contributions in this paper:


\begin{itemize}
    \item We release XL-Sum, a dataset containing 1 million article-summary pairs in 44 languages, being the first publicly available abstractive summarization dataset for many of them.
    
    \item We create a data curation tool that can automatically crawl and extract article-summary pairs from BBC, using which the dataset can be made even larger over time.
    
    \item We are the first to perform multilingual summarization on a diverse set of languages, achieving strong baselines on all languages tested.

    
\end{itemize}

We are releasing the dataset, curation tool, and summarization model checkpoints. We believe that our efforts in this work will encourage the community to push the boundaries of abstractive text summarization beyond the English language, especially for low and mid-resource languages.

\section{The XL-Sum Dataset}

In this section, we present details of the XL-Sum dataset together with the curation process.
Table \ref{tab:dataset} shows the article-summary statistics for all languages in the XL-Sum dataset. 

\subsection{Content Source}

BBC publishes news in 43 languages\footnote{\url{https://www.bbc.co.uk/ws/languages}} ranging from low-resource languages such as Bengali and Swahili to high-resource ones including English and Russian. Among the 43 languages, Serbian is a special case that is published in both Cyrillic, the official script, and Latin, the colloquial script. We treat them as different languages in this work, totaling to a coverage of 44 languages.

\subsection{Content Search}

As BBC does not provide any archive or RSS feed on their website, we designed a crawler to recursively crawl pages starting from the homepage by visiting different article links present in each page visited. We were able to take advantage of the fact that all BBC sites have somewhat similar structures, and were able to scrape articles from all sites. 
We discarded pages with no textual contents (mostly pages consisting of multimedia contents) before further processing. 

\subsection{Article-Summary Extraction}

The process of automatically collecting summaries of articles differs across different datasets. For example, the CNN/DM dataset \citep{hermann2015teaching} merged bullet point highlights provided with the articles as reference summaries, whereas the XSum dataset \citep{narayan2018don} used the first line of the article as the summary and the rest of the article as the input.

Our method of collecting summaries was made easier by the consistent editorial style of the BBC articles we crawled. BBC typically provides a summary of a whole article in the form of a bold paragraph containing one or two sentences at the beginning of each article. These summaries are written professionally by the authors of the articles in order to convey its main story within one small paragraph. This is in contrast to the headline which serves to draw the attention of viewers into reading the article. (We show an example article-summary pair from BBC English in Table \ref{tab:example}.) We designed a number of heuristics to make the extraction effective by carefully examining the HTML structures of the crawled pages:
\begin{enumerate}\itemsep0em
    \item The desired summary must be present within the beginning two paragraphs of an article.
    \item The summary paragraph must have some portion of texts in bold format.
    \item The summary paragraph may contain some hyperlinks that may not be bold. The proportion of bold texts and hyperlinked texts to the total length of the paragraph in consideration must be at least 95\%.
    \item All texts except the summary and the headline must be included in the input text (including image captions).
    \item The input text must be at least twice as large as the summary.
\end{enumerate}

Any sample that did not conform to these heuristics were discarded. Our strategy of automatic annotation of summaries resembles XSum to some extent, but we found the first line to contain meta-information in many articles (e.g., author information, date of last modification). As such, we used the bold paragraphs as the summaries instead.



\section{Human Evaluation of XL-Sum}\label{sec:human}


Despite being written by professionals, evaluating the quality of the XL-Sum dataset is crucial to ensure that it can be valuable and used by a wider community for abstractive summarization. For this, we thoroughly conducted human evaluations on a subset of the dataset. 





We hired professional annotators\footnote{Each evaluator had at least an undergraduate degree and had native or bilingual proficiency in the languages they were assigned to. Bilingual proficiency was determined by professional language tests for the corresponding language.} to assess the quality of the top-10 languages according to the number of speakers worldwide.\footnote{\url{https://w.wiki/Pss}} It is worth noting that not all of these 10 languages are high-resource (e.g., Bengali is a low-resource language despite being one of the most widely spoken). 

Each evaluator was asked to assess the quality of a random subset of the dataset (around 250 article-summary pairs) by answering the following questions with `Yes'/`No':



\begin{description}
    \item[Property A:] Does the summary convey what the article is about?
    \item[Property B:] If the answer to property \verb!A! is `Yes', does the summary contain any information that is inconsistent with the article?
    \item[Property C:] If the answer to property \verb!A! is `Yes', does the summary contain any information that cannot be inferred from the article?
\end{description}

The motivation for designing these properties stemmed from recent progress on the quality estimation of neural language generation (NLG) models. \citet{belinkov2017synthetic} showed that NLG models are vulnerable to noisy and low-quality training samples, hence, it is essential to validate the quality of the summary through Property \verb!A!. Ensuring factual consistency and faithfulness \citep{wang2020asking, maynez2020faithfulness} of the generated summaries is also a key consideration for neural abstractive summarization since neural models have been shown to generate hallucinated texts \citep{see-etal-2017-get}. Property \verb!B! checks for consistency between the article and the summary; while Property \verb!C! assesses the hallucination implicitly by limiting the knowledge domain to the input article and identifying additional information present in the summary.

\begin{table}[h]
    \centering
    \begin{tabular}{l}
    \hline
    \textbf{Language/Dataset}\\
    \hline
    CNN/DM\\
    XSum\\
    \hline
    English\\
    Chinese\\
    Hindi\\
    Spanish\\
    French\\
    Arabic\\
    Bengali\\
    Russian\\
    Portuguese\\
    Indonesian\\
    \hline
    \end{tabular}
    \begin{tabular}{ccc}
    \hline
    \textbf{A} & \textbf{B} & \textbf{C}\\
    \hline
    98.33 & 1.22 & 24.57\\
    92.00 & 0.00 & 71.74\\
    \hline
    99.66 & 0.00 & 37.37\\
    93.49 & 0.00 & 29.56\\
    90.91 & 0.00 & 31.42\\
    84.71 & 0.00 & 42.93\\
    99.20 & 0.00 & 26.72\\
    98.34 & 0.00 & 25.31\\
    91.14 & 0.00 & 26.85\\
    95.65 & 0.00 & 38.64\\
    88.31 & 0.47 & 38.50\\
    97.59 & 0.41 & 27.57\\
    \hline
    \end{tabular}
    \caption{Quality of XL-Sum (along with CNN/DM and XSum) assessed by human evaluators. In most of the cases, the evaluators agree that the summaries convey the main idea (Property A), and do not conflict with the input text (Property B). However, the summaries may contain some additional information (Property C), since the editors writing the summaries may use their common sense and additional domain knowledge not present in the article.}\label{tab:human}
\end{table}

It is desirable that property \verb!A! would have a high percentage of `Yes', while both property \verb!B! and \verb!C! would have a low percentage. However, the third property can be a bit ambiguous, since some pieces of information may not be directly present in the input article, but a professional with background knowledge and understanding of the article topic may infer them inherently. Since text summarization is posed as a closed-domain task, we asked the annotators to label accordingly, i.e., not making use of any additional information outside the article. We provided them with some sample annotations in English to aid them during annotation. The percentages of `Yes' for the three properties are shown in Table \ref{tab:human}.\footnote{We measured inter-annotator agreement using Cohen's kappa coefficient. Most scores were within the 0.7-0.9 range, showing high agreement between the evaluators} We also showed human evaluation of CNN/DM and XSum for contrast. Each article-summary pair was labeled by two separate evaluators, and we considered each pair to exhibit property \verb!A! if both agreed upon it and property \verb!B! and \verb!C! if at least one of them did.

We can see from Table \ref{tab:human} (Property \verb!A!) that most languages showed a high percentage of good summaries in the upper nineties, while some had a slightly lower percentage (i.e., Spanish and Portuguese). We were informed by the annotators that the negative summaries were mostly extracted from opinion pieces and blog posts, the bold paragraphs of which did not convey the main story of the articles. 

Almost none of the summaries contained any conflicting information (Property \verb!B!), while about one-third on average had information that was not directly inferrable (Property \verb!C!). Interestingly, more than 75\% of the latter had missing information such as first names, designations, or elaborations of acronyms. For example, a summary had the phrase `President-elect Joe Biden' while its corresponding article had no presence of the first name `Joe'. Another article had NHS elaborated in the summary, which was absent in the article. In general. the types of extra information carried in the summaries were more or less the same across all languages. CNN/DM and XSum exhibit extra information in the summaries as well, implying this phenomenon is common in abstractive text summarization datasets.

The presence of extra information in the summaries is understandable, since professional experts writing these summaries not only use the information present in the article text, but also incorporate their knowledge and understanding of the outside world. But for a closed-domain summarization model or a layman to the topic, inferring this information is not so straightforward, which makes the automatic abstractive summarization task more challenging.
This phenomenon may explain why language models fine-tuned on pretrained checkpoints \citep{raffel2019exploring, yan2020prophetnet, zhang2020pegasus} are achieving state-of-the-art results in abstractive summarization, as they are able to make use of outside information gained from the high volume of texts they were pretrained with. Additionally, it would be interesting to investigate whether the recent trend of incorporating real-world knowledge and commonsense reasoning~\cite{tandon2018commonsense, deng2020integrating} into language models could improve text summarization performance.

\begin{table*}[h]
    \centering
    \begin{tabular}{l}
    \hline
    \textbf{Language}\\
    \textbf{/Dataset}\\
    \hline
    CNN/DM\\
    XSum\\
    \hline
    English\\
    Chinese\\
    Hindi\\
    Spanish\\
    French\\
    Arabic\\
    Bengali\\
    Russian\\
    Portuguese\\
    Indonesian\\
    \hline
    \end{tabular}
    \begin{tabular}{c c c c}
    \hline
    \multicolumn{4}{c}{\textbf{Percentage of novel n-grams} $\uparrow$}\\
    \hline
    \textbf{n = 1} & \textbf{n = 2} & \textbf{n = 3} & \textbf{n = 4}\\
    \hline
    13.20 & 52.77 & 72.22 & 81.40\\
    35.76 & 83.45 & 95.50 & 98.49\\
    \hline
    32.22 & 80.99 & 94.57 & 98.06\\
    36.13 & 79.23 & 91.14 & 94.58\\
    29.55 & 74.77 & 90.87 & 96.29\\
    32.63 & 76.29 & 91.96 & 96.57\\
    35.41 & 74.72 & 88.39 & 93.24\\
    49.88 & 84.56 & 94.79 & 98.10\\
    38.81 & 81.10 & 92.10 & 95.89\\
    49.27 & 85.89 & 95.57 & 98.34\\
    30.28 & 77.11 & 92.23 & 96.71\\
    33.54 & 76.87 & 91.73 & 96.53\\
    \hline
    \end{tabular}
    \begin{tabular}{c c c c}
    \hline
    \multirow{2}{*}{\textbf{ABS $\big\uparrow$}} & \multirow{2}{*}{\textbf{CMP $\big\uparrow$}} & \multirow{2}{*}{\textbf{RED (n=1) $\big\downarrow$}} & \multirow{2}{*}{\textbf{RED (n=2) $\big\downarrow$}}\\
    \\
    \hline
    38.75 & 90.90 & 13.73 & 1.10\\
    75.70 & 90.40 & 5.83 & 0.16\\
    \hline
    71.74 & 92.97 & 6.56 & 0.20\\
    70.23 & 92.95 & 7.37 & 0.50\\
    64.63 & 93.00 & 9.91 & 0.16\\
    66.60 & 92.49 & 11.45 & 0.57\\
    65.29 & 88.34 & 8.34 & 0.44\\
    76.72 & 90.62 & 3.93 & 0.18\\
    72.76 & 94.74 & 2.93 & 0.25\\
    78.39 & 91.25 & 4.34 & 0.16\\
    66.80 & 94.47 & 10.22 & 0.34\\
    66.68 & 91.62 & 3.94 & 0.23\\
    \hline
    \end{tabular}
    \caption{Intrinsic evaluation of our XL-Sum dataset compared to CNN/Daily Mail and XSum. All values are reported in percentage for easier comparison. We use $\protect\uparrow$ to indicate ``higher is better'' and $\protect\downarrow$ for the reverse. Both of XL-Sum and XSum are highly abstractive, concise, and shows comparable quality, although the XSum dataset contains only English samples. For both XL-Sum and XSum, percentages of novel n-grams (n = 1, 2, 3, 4) are significantly higher than CNN/DM. High abstractiveness (ABS) scores of XL-Sum and XSum also bolster this finding. Additionally, low redundancy (RED) and high compression (CMP) values indicate that XL-Sum and XSum are more concise than CNN/DM.}\label{tab:intrinsic}
    \end{table*}

\section{Intrinsic Evaluation of XL-Sum}

Although the human evaluations provided good insights into the quality of the summaries, there are several other aspects of the summaries that are often infeasible or impractical to judge by human evaluators. With the above backdrop, several works \citep{narayan2018don, grusky2018newsroom, bommasani2020intrinsic} have proposed many automatic metrics to quantify important features of abstractive summaries (e.g., novel words, abstractivity, compression, and redundancy).  


\textbf{Novel n-gram ratio:} \citet{narayan2018don} proposed the percentage of n-grams in the summary that do not occur in the input article as a means of measuring abstractiveness.  

\textbf{Abstractivity:} \citet{grusky2018newsroom} introduced \textit{fragments}, which greedily match text spans between the article and the summary, and \citet{bommasani2020intrinsic} generalized it to introduce \textit{abstractivity} to measure absractiveness. 

\textbf{Compression:} \citet{bommasani2020intrinsic} proposed \textit{compression} as a metric to quantify conciseness. Compression is measured by \begin{equation}
    \textbf{CMP}(A, S) = 1 - \frac{|S|}{|A|}
\end{equation} 
where $|A|$ and $|S|$ denote the length of the article and the summary, respectively. We measured length in terms of number of tokens.

\textbf{Redundancy:} Although \citet{bommasani2020intrinsic} proposed a metric to measure \textit{redundancy}, it is only applicable to multi-sentence summaries, which is not the case for most examples in XL-Sum. Thus, we propose a new redundancy metric by calculating the number of repetitive n-grams in the summary text. 

Let $\{g_1, g_2, \cdots , g_m\}$ be the unique n-grams occurring in a summary $S$, and let $\{f_1, f_2, \cdots, f_m\}$ be their respective frequencies. Then the total number of repeated n-grams are $\sum_{i=1}^{m} (f_i - 1)$. We define redundancy as the ratio of redundant n-grams and the total number of n-grams in $S$: 

\begin{align*}
    \textbf{RED}(S) & = \frac{\sum_{i=1}^{m} (f_i - 1)}{\sum_{i=1}^{m} f_i}\\
    & = 1 - \frac{m}{|S| - n + 1} \numberthis 
\end{align*}

It is preferred of a good summary to have a high novel n-gram ratio, abstractivity, and compression; while having a low redundancy score. We show these metrics in Table \ref{tab:intrinsic} (for redundancy, we report values for $n = 1, 2$). 
We also show these metrics for the CNN/DM and XSum datasets.

The results indicate that the XL-Sum dataset is highly abstractive---about one-third of the tokens, and more than 75\% of the bigrams in the summaries are novel, and the abstractiveness score is also high (more than 65\% for most of the languages). Additionally, XL-Sum is very concise (the summary is less than one-tenth of the input article for most of the languages), and contains minimal redundancy (less than 10\% for the majority). The quality of XSum is comparable, however, it is limited to only one language (i.e., English). On the other hand, the CNN/Daily Mail dataset is significantly behind XL-Sum and XSum as indicated by most of the metrics mentioned above.


\section{Experiments and Benchmarks}

In previous sections, we have discussed the quality of XL-Sum. In addition, it is imperative to see how state-of-the-art models perform when trained on this dataset. Moreover, for many languages (e.g., Bengali, Swahili), currently, there is no publicly available dataset and benchmarks for abstractive text summarization to the best of our knowledge. In this section, we train summarization models with the XL-Sum dataset and provide several baselines and benchmark results.

Fine-tuning Transformer-based \citep{vaswani2017attention} seq2seq models initialized with pretrained weights from self-supervised training \citep{raffel2019exploring, liu2019text,  rothe2020leveraging, yan2020prophetnet, zhang2020pegasus} has been shown to achieve state-of-the-art performance on many abstractive text summarization datasets. There are many multilingual pretrained checkpoints available through the Hugging Face Transformers Library \citep{wolf2020transformers}. Among them, we chose to use the mT5 model \citep{xue2020mt5}, a multilingual language model pretrained on a large dataset of 101 languages.
  

We performed summarization experiments in two settings: (i) multilingual, and (ii) low-resource. 
For performance reporting, for each language, we randomly sampled 500 pairs for its development set and 500 pairs for its test set, while using the rest of the pairs for training. We tokenized our training samples using the 250k wordpiece \citep{wu2016google} vocabulary provided with the mT5 checkpoint. Due to computational constraints, we used the base model (600M parameters) and had to truncate the inputs to 512 tokens and the outputs to 64 tokens. We used the ROUGE-1, ROUGE-2 and ROUGE-L \citep{lin2004rouge} scores for automatic evaluation. For inference, we used beam search with beam size 4 and length penalty of $\alpha=0.6$ \citep{wu2016google}. 

\subsection{Multilingual Summarization}

Multilingual training is performed by training a single model with training samples from multiple languages. It has been previously used in several NLP tasks including neural machine translation \citep{arivazhagan2019massively} and language model pretraining \citep{lample2019cross}. However, multilingual training in the context of abstractive summarization has not been a major focus of the community. As such, the aim of this experiment is to demonstrate that a single model can perform well on summarizing texts in different languages, and that sister languages with morphological similarity can take advantage of positive transfer from each other which is not possible in monolingual settings.

For this experiment, we followed a similar training strategy as \citet{lample2019cross}: we sampled each batch from a single language containing 256 samples and used a smoothing factor ($\alpha$) of 0.5 
so that batches of low-resource languages would be sampled at a higher rate, increasing their frequency during training.

We fine-tuned the mT5 model for 35k steps on a distributed cluster of 8 Nvidia Tesla P100 GPUs for 4 days. We used the Adafactor optimizer \citep{shazeer2018adafactor} with a linear warmup of 5,000 steps and `inverse square root'  learning rate schedule. We show the ROUGE scores achieved by the model on the top-10 languages in Table \ref{tab:mult}.

\begin{table}[h]
\centering
\begin{tabular}{l}
\hline
\textbf{Language}\\
\hline
English\\
Chinese\\
Hindi\\
Spanish\\
French\\
Arabic\\
Bengali\\
Russian\\
Portuguese\\
Indonesian\\
\hline
\end{tabular}
\begin{tabular}{ccc}
\hline
\textbf{R-1} & \textbf{R-2} & \textbf{R-L}\\
\hline
36.99 & 15.18 & 29.64\\
36.89 & 15.23 & 30.52\\
34.51 & 13.55 & 28.23\\
30.93 & 12.14 & 23.76\\
34.47 & 15.94 & 27.53\\
33.23 & 13.74 & 27.84\\
28.32 & 11.43 & 24.23\\
31.10 & 13.47 & 25.54\\
31.06 & 11.62 & 23.39\\
36.17 & 16.70 & 30.50\\
\hline
\end{tabular}
\caption{ROUGE scores for multilingual summarization achieved by the mT5 model when fine-tuned on the XL-Sum training set.}\label{tab:mult}
\end{table}

As we can see from the table, the multilingual model achieved higher than 11 ROUGE-2 scores on all languages. Some of these languages (e.g., Bengali) are low-resource, but the model still obtained competitive results comparable to high and mid-resource languages. Also, we are the first to report the abstractive summarization benchmark for a number of languages, including Bengali.

The mT5-base model achieves a R2-score of 15.18 on the English language. 
In comparison, the state-of-the-art PEGASUS$_\text{BASE}$ model~\cite{zhang2020pegasus} obtained an R-2 score of 16.58 trained on the XSum English dataset, which is similar to XL-Sum in nature. This result suggests that the performance is comparable to the state-of-the-art English summarization. The R-2 scores for other languages are also similar to English, indicating that our dataset can help to effectively generate automatic summaries for all languages tested, including those low-resource ones.

\subsection{Low-Resource Summarization}


We have shown the effectiveness of the multilingual training strategy in summarizing articles for a wide set of languages with a single model. However, training the model is compute-heavy and it may not be realistic in many scenarios. For the dataset to be termed as versatile, models trained on individual languages should be able to perform competitively with the multilingual model. To confirm this is indeed the case, we performed training on five low-resource languages from Table \ref{tab:dataset} (Amharic, Azerbaijani, Bengali, Japanese, Swahili) in a compute-efficient setup. We fine-tuned mT5 on each language separately for 6-10 epochs (since the total training samples were limited, we had to be careful to prevent overfitting) on a single-GPU (Nvidia RTX 2080Ti) machine. For these experiments, we used a batch size of 32 and trained with a slanted learning rate schedule \citep{howard2018universal}. We show the ROUGE scores of each model in Table \ref{tab:low}. We use the results from the multilingual models as baseline. 

\begin{table}[h]
\centering
\begin{tabular}{l}
\hline
\multirow{2}{*}{\textbf{Language}}\\
\\
\hline
Amharic\\
Azerbaijani\\
Bengali\\
Japanese\\
Swahili\\
\hline
\end{tabular}
\begin{tabular}{c @{\hspace{0.2cm}} c}
\hline
\textbf{Low-resource} & \textbf{Multilingual}\\
R-1/R-2/R-L & R-1/R-2/R-L\\
\hline
\small{15.33/5.12/13.85} & \small{\textbf{17.49}/\textbf{6.34}/\textbf{15.76}}\\ 
\small{16.79/6.94/15.36} & \small{\textbf{19.29}/\textbf{8.20}/\textbf{17.62}}\\
\small{25.33/9.50/22.02} & \small{\textbf{28.32}/\textbf{11.43}/\textbf{24.02}}\\
\small{44.55/21.35/34.43} & \small{\textbf{47.17}/\textbf{23.34}/\textbf{36.20}}\\
\small{34.29/15.97/28.21} & \small{\textbf{38.18}/\textbf{18.16}/\textbf{30.98}}\\
\hline
\end{tabular}
\caption{Performance of mT5 model fine-tuned on a low-resource training setup vs multi-lingual setup as mentioned in the previous section.}\label{tab:low}
\end{table}

As evident by the results from Table \ref{tab:low}, the multilingual model outperformed all the models trained on a single language. This is expected since similar languages can have positive transfer between them \citep{conneau2019unsupervised} when trained together. However, the low-resource models do not trail by a large margin, in all cases the difference is not more than 2 for R-2 scores. This is a good indication that models fine-tuned on such a low amount of samples can still generalize to produce results competitive to multilingual models.

The case for Amharic, Azerbaijani and Japanese call for a discussion on their performance. The first two had comparatively low scores, while the last one (Japanese) had considerably higher scores compared to the other languages. Amharic and Azerbaijani had approximately 4k and 6k training samples respectively, which we conjecture is the primary reason behind their underperformance. Moreover, we did not find any reliable stemmer to preprocess the generated summaries before computing ROUGE, which may also hurt the scores. On the other hand, Japanese texts are not word-segmented and the words need to be separated before calculating ROUGE. We used Fugashi \citep{mccann2020fugashi}, and possibly due to its aggressive segmentation, the scores turned out to be higher than other languages. Similar high results have also been reported while measuring BLEU \citep{papineni2002bleu} scores for machine translation evaluation in Japanese \citep{kudo2018subword}. 

Results in Table~\ref{tab:low} show that although these languages are low-resource, the scores of the two setups are close, indicating our dataset can also be useful when used with a constrained computation capability. This is likely to contribute towards advances in low-resource text summarization, enabling fairness and access to the under-served communities.

\section{Related Works}
\citet{rush2015neural, nallapati2016abstractive} pioneered neural abstractive summarization, using recurrent attentional seq2seq models \citep{bahdanau2014neural}. \citet{see-etal-2017-get} introduced Pointer-Generator networks for abstractive summarization, which can learn to copy words from the input text, in addition to generating new texts with the decoder. \citet{gehring2017convolutional} proposed convolutional seq2seq models and applied it to perform abstractive summarization. \citet{narayan2018don} extended the work by integrating topic embeddings with the model.

Pretrained language models have recently been successfully applied to abstractive summarization. \citet{liu2019text} initialized the encoder and \citet{rothe2020leveraging} initialized both the encoder and the decoder of a seq2seq model with the pre-trained BERT \citep{devlin2018bert} weights and fine-tuned the models for abstractive summarization. \citet{raffel2019exploring, yan2020prophetnet} used fully pre-trained seq2seq models, while \citet{zhang2020pegasus} introduced a summarization-specific pre-training objective to achieve state-of-the-art results on multiple datasets. 

Most works on abstractive summarization have so far focused on English, in large part due to a lack of benchmark datasets for other languages. \citet{giannakopoulos2015multiling} introduced MultiLing 2015, a summarization dataset spanning 40 languages. However, MultiLing 2015 is limited in size, with the training set having only 10k samples in total. \citet{cao2020multisumm, scialom2020mlsum} introduced two new datasets for multilingual summarization, but both were limited to less than 10 languages. Moreover, samples for different languages were collected from different sources, making them exposed to different types of summarization strategies, which raises questions on the uniformity of the summaries.


\section{Conclusion and Future Works}

In this paper, we present XL-Sum, a large-scale, high-quality multilingual text summarization dataset, containing 1 million samples across 44 languages collected from a single source, BBC. For many of the languages, XL-Sum provides the first publicly available abstractive summarization dataset and benchmarks. We also make the dataset curation tool available for the researchers, which will help to grow the dataset over time. Thorough human and intrinsic evaluations indicate that the summaries in our dataset are highly abstractive and concise while conveying the main idea with little to no conflict with the input article. Additionally, we demonstrate that multilingual training can help towards better summarization, most likely due to the positive transfer between sister languages with morphological similarity. Moreover, XL-Sum is also useful in a low-resource and compute-efficient setting.

In future, we will investigate the use of our dataset for other summarization tasks (e.g., cross-lingual summarization \citealp{zhu2019ncls}). 

We hope the XL-Sum dataset will be helpful for the research community, especially for the researchers working to ensure fair access to technological advances to under-served communities with low-resource languages.

\section*{Acknowledgements}
 
This work was performed using the OzSTAR national facility at the Swinburne University of Technology. The OzSTAR program receives funding in part from the Astronomy National Collaborative Research Infrastructure Strategy (NCRIS) allocation provided by the Australian Government.





\bibliographystyle{acl_natbib}
\bibliography{anthology,acl2021}

\begin{thebibliography}{44}
\expandafter\ifx\csname natexlab\endcsname\relax\def\natexlab#1{#1}\fi

\bibitem[{Arivazhagan et~al.(2019)Arivazhagan, Bapna, Firat, Lepikhin, Johnson,
  Krikun, Chen, Cao, Foster, Cherry et~al.}]{arivazhagan2019massively}
Naveen Arivazhagan, Ankur Bapna, Orhan Firat, Dmitry Lepikhin, Melvin Johnson,
  Maxim Krikun, Mia~Xu Chen, Yuan Cao, George Foster, Colin Cherry, et~al.
  2019.
\newblock \href {https://arxiv.org/abs/1907.05019} {Massively multilingual
  neural machine translation in the wild: Findings and challenges}.
\newblock \emph{arXiv preprint arXiv:1907.05019}.

\bibitem[{Bahdanau et~al.(2015)Bahdanau, Cho, and Bengio}]{bahdanau2014neural}
Dzmitry Bahdanau, Kyunghyun Cho, and Yoshua Bengio. 2015.
\newblock \href {http://arxiv.org/abs/1409.0473} {Neural machine translation by
  jointly learning to align and translate}.
\newblock In \emph{Proceedings of the 3rd International Conference on Learning
  Representations (ICLR 2015)}, San Diego, California, USA.

\bibitem[{Belinkov and Bisk(2018)}]{belinkov2017synthetic}
Yonatan Belinkov and Yonatan Bisk. 2018.
\newblock \href {https://arxiv.org/abs/1711.02173} {Synthetic and natural noise
  both break neural machine translation}.
\newblock In \emph{Proceedings of the 6th International Conference on Learning
  Representations (ICLR 2018)}, Vancouver, BC, Canada.

\bibitem[{Bommasani and Cardie(2020)}]{bommasani2020intrinsic}
Rishi Bommasani and Claire Cardie. 2020.
\newblock \href {https://doi.org/10.18653/v1/2020.emnlp-main.649} {Intrinsic
  evaluation of summarization datasets}.
\newblock In \emph{Proceedings of the 2020 Conference on Empirical Methods in
  Natural Language Processing (EMNLP)}, pages 8075--8096, Online. Association
  for Computational Linguistics.

\bibitem[{Cao et~al.(2020)Cao, Wan, Yao, and Yu}]{cao2020multisumm}
Yue Cao, Xiaojun Wan, Jinge Yao, and Dian Yu. 2020.
\newblock \href {https://aaai.org/ojs/index.php/AAAI/article/view/5328}
  {Multisumm: Towards a unified model for multi-lingual abstractive
  summarization}.
\newblock In \emph{Proceedings of Thirty-Fourth {AAAI} Conference on Artificial
  Intelligence, {AAAI} 2020}, pages 11--18. {AAAI} Press.

\bibitem[{Cho et~al.(2014)Cho, van Merri{\"e}nboer, Gulcehre, Bahdanau,
  Bougares, Schwenk, and Bengio}]{cho2014learning}
Kyunghyun Cho, Bart van Merri{\"e}nboer, Caglar Gulcehre, Dzmitry Bahdanau,
  Fethi Bougares, Holger Schwenk, and Yoshua Bengio. 2014.
\newblock \href {https://doi.org/10.3115/v1/D14-1179} {Learning phrase
  representations using {RNN} encoder{--}decoder for statistical machine
  translation}.
\newblock In \emph{Proceedings of the 2014 Conference on Empirical Methods in
  Natural Language Processing ({EMNLP})}, pages 1724--1734, Doha, Qatar.
  Association for Computational Linguistics.

\bibitem[{Cohn and Lapata(2008)}]{cohn2008sentence}
Trevor Cohn and Mirella Lapata. 2008.
\newblock \href {https://www.aclweb.org/anthology/C08-1018} {Sentence
  compression beyond word deletion}.
\newblock In \emph{Proceedings of the 22nd International Conference on
  Computational Linguistics (Coling 2008)}, pages 137--144, Manchester, UK.

\bibitem[{Conneau et~al.(2020)Conneau, Khandelwal, Goyal, Chaudhary, Wenzek,
  Guzm{\'a}n, Grave, Ott, Zettlemoyer, and Stoyanov}]{conneau2019unsupervised}
Alexis Conneau, Kartikay Khandelwal, Naman Goyal, Vishrav Chaudhary, Guillaume
  Wenzek, Francisco Guzm{\'a}n, Edouard Grave, Myle Ott, Luke Zettlemoyer, and
  Veselin Stoyanov. 2020.
\newblock \href {https://doi.org/10.18653/v1/2020.acl-main.747} {Unsupervised
  cross-lingual representation learning at scale}.
\newblock In \emph{Proceedings of the 58th Annual Meeting of the Association
  for Computational Linguistics}, pages 8440--8451, Online. Association for
  Computational Linguistics.

\bibitem[{Conneau and Lample(2019)}]{lample2019cross}
Alexis Conneau and Guillaume Lample. 2019.
\newblock \href
  {https://papers.nips.cc/paper/2019/file/c04c19c2c2474dbf5f7ac4372c5b9af1-Paper.pdf}
  {Cross-lingual language model pretraining}.
\newblock In \emph{Proceedings of the 33rd International Conference on Neural
  Information Processing Systems (NeurIPS 2019)}, pages 7059--7069, Vancouver,
  Canada.

\bibitem[{Deng et~al.(2020)Deng, Ji, Rainey, Zhang, and
  Lu}]{deng2020integrating}
Changyu Deng, Xunbi Ji, Colton Rainey, Jianyu Zhang, and Wei Lu. 2020.
\newblock \href {https://doi.org/10.1016/j.isci.2020.101656} {Integrating
  machine learning with human knowledge}.
\newblock \emph{iScience}, 23(11):101656.

\bibitem[{Devlin et~al.(2019)Devlin, Chang, Lee, and
  Toutanova}]{devlin2018bert}
Jacob Devlin, Ming-Wei Chang, Kenton Lee, and Kristina Toutanova. 2019.
\newblock \href {https://www.aclweb.org/anthology/N19-1423} {{BERT}:
  Pre-training of deep bidirectional transformers for language understanding}.
\newblock In \emph{Proceedings of the 2019 Conference of the North {A}merican
  Chapter of the Association for Computational Linguistics: Human Language
  Technologies, Volume 1 (Long and Short Papers)}, pages 4171--4186,
  Minneapolis, Minnesota, USA. Association for Computational Linguistics.

\bibitem[{Gehring et~al.(2017)Gehring, Auli, Grangier, Yarats, and
  Dauphin}]{gehring2017convolutional}
Jonas Gehring, Michael Auli, David Grangier, Denis Yarats, and Yann~N Dauphin.
  2017.
\newblock \href {http://proceedings.mlr.press/v70/gehring17a.html}
  {Convolutional sequence to sequence learning}.
\newblock In \emph{Proceedings of the 34th International Conference on Machine
  Learning, {ICML} 2017}, volume~70, pages 1243--1252. {PMLR}.

\bibitem[{Giannakopoulos et~al.(2015)Giannakopoulos, Kubina, Conroy,
  Steinberger, Favre, Kabadjov, Kruschwitz, and
  Poesio}]{giannakopoulos2015multiling}
George Giannakopoulos, Jeff Kubina, John Conroy, Josef Steinberger, Benoit
  Favre, Mijail Kabadjov, Udo Kruschwitz, and Massimo Poesio. 2015.
\newblock \href {https://doi.org/10.18653/v1/W15-4638} {{M}ulti{L}ing 2015:
  Multilingual summarization of single and multi-documents, on-line fora, and
  call-center conversations}.
\newblock In \emph{Proceedings of the 16th Annual Meeting of the Special
  Interest Group on Discourse and Dialogue}, pages 270--274, Prague, Czech
  Republic. Association for Computational Linguistics.

\bibitem[{Grusky et~al.(2018)Grusky, Naaman, and Artzi}]{grusky2018newsroom}
Max Grusky, Mor Naaman, and Yoav Artzi. 2018.
\newblock \href {https://doi.org/10.18653/v1/N18-1065} {{N}ewsroom: A dataset
  of 1.3 million summaries with diverse extractive strategies}.
\newblock In \emph{Proceedings of the 2018 Conference of the North {A}merican
  Chapter of the Association for Computational Linguistics: Human Language
  Technologies, Volume 1 (Long Papers)}, pages 708--719, New Orleans,
  Louisiana. Association for Computational Linguistics.

\bibitem[{Hermann et~al.(2015)Hermann, Ko\v{c}isk\'{y}, Grefenstette, Espeholt,
  Kay, Suleyman, and Blunsom}]{hermann2015teaching}
Karl~Moritz Hermann, Tom\'{a}\v{s} Ko\v{c}isk\'{y}, Edward Grefenstette, Lasse
  Espeholt, Will Kay, Mustafa Suleyman, and Phil Blunsom. 2015.
\newblock \href
  {https://papers.nips.cc/paper/2015/file/afdec7005cc9f14302cd0474fd0f3c96-Paper.pdf}
  {Teaching machines to read and comprehend}.
\newblock In \emph{Proceedings of the 28th International Conference on Neural
  Information Processing Systems (NIPS 2015)}, pages 1693--1701, Montreal,
  Canada.

\bibitem[{Howard and Ruder(2018)}]{howard2018universal}
Jeremy Howard and Sebastian Ruder. 2018.
\newblock \href {https://doi.org/10.18653/v1/P18-1031} {Universal language
  model fine-tuning for text classification}.
\newblock In \emph{Proceedings of the 56th Annual Meeting of the Association
  for Computational Linguistics (Volume 1: Long Papers)}, pages 328--339,
  Melbourne, Australia. Association for Computational Linguistics.

\bibitem[{Hsu et~al.(2018)Hsu, Lin, Lee, Min, Tang, and Sun}]{hsu2018unified}
Wan-Ting Hsu, Chieh-Kai Lin, Ming-Ying Lee, Kerui Min, Jing Tang, and Min Sun.
  2018.
\newblock \href {https://doi.org/10.18653/v1/P18-1013} {A unified model for
  extractive and abstractive summarization using inconsistency loss}.
\newblock In \emph{Proceedings of the 56th Annual Meeting of the Association
  for Computational Linguistics (Volume 1: Long Papers)}, pages 132--141,
  Melbourne, Australia. Association for Computational Linguistics.

\bibitem[{Kudo(2018)}]{kudo2018subword}
Taku Kudo. 2018.
\newblock \href {https://doi.org/10.18653/v1/P18-1007} {Subword regularization:
  Improving neural network translation models with multiple subword
  candidates}.
\newblock In \emph{Proceedings of the 56th Annual Meeting of the Association
  for Computational Linguistics (Volume 1: Long Papers)}, pages 66--75,
  Melbourne, Australia. Association for Computational Linguistics.

\bibitem[{Lin(2004)}]{lin2004rouge}
Chin-Yew Lin. 2004.
\newblock \href {https://www.aclweb.org/anthology/W04-1013} {{ROUGE}: A package
  for automatic evaluation of summaries}.
\newblock In \emph{Text Summarization Branches Out}, pages 74--81, Barcelona,
  Spain. Association for Computational Linguistics.

\bibitem[{Liu and Lapata(2019)}]{liu2019text}
Yang Liu and Mirella Lapata. 2019.
\newblock \href {https://doi.org/10.18653/v1/D19-1387} {Text summarization with
  pretrained encoders}.
\newblock In \emph{Proceedings of the 2019 Conference on Empirical Methods in
  Natural Language Processing and the 9th International Joint Conference on
  Natural Language Processing (EMNLP-IJCNLP)}, pages 3730--3740, Hong Kong,
  China. Association for Computational Linguistics.

\bibitem[{Maynez et~al.(2020)Maynez, Narayan, Bohnet, and
  McDonald}]{maynez2020faithfulness}
Joshua Maynez, Shashi Narayan, Bernd Bohnet, and Ryan McDonald. 2020.
\newblock \href {https://doi.org/10.18653/v1/2020.acl-main.173} {On
  faithfulness and factuality in abstractive summarization}.
\newblock In \emph{Proceedings of the 58th Annual Meeting of the Association
  for Computational Linguistics}, pages 1906--1919, Online. Association for
  Computational Linguistics.

\bibitem[{McCann(2020)}]{mccann2020fugashi}
Paul McCann. 2020.
\newblock \href {https://doi.org/10.18653/v1/2020.nlposs-1.7} {fugashi, a tool
  for tokenizing {J}apanese in python}.
\newblock In \emph{Proceedings of Second Workshop for NLP Open Source Software
  (NLP-OSS)}, pages 44--51, Online. Association for Computational Linguistics.

\bibitem[{Nallapati et~al.(2016)Nallapati, Zhou, dos Santos,
  GuÌ‡l{\c{c}}ehre, and Xiang}]{nallapati2016abstractive}
Ramesh Nallapati, Bowen Zhou, Cicero dos Santos, {\c{C}}a{\u{g}}lar
  GuÌ‡l{\c{c}}ehre, and Bing Xiang. 2016.
\newblock \href {https://doi.org/10.18653/v1/K16-1028} {Abstractive text
  summarization using sequence-to-sequence {RNN}s and beyond}.
\newblock In \emph{Proceedings of The 20th {SIGNLL} Conference on Computational
  Natural Language Learning}, pages 280--290, Berlin, Germany. Association for
  Computational Linguistics.

\bibitem[{Narayan et~al.(2018)Narayan, Cohen, and Lapata}]{narayan2018don}
Shashi Narayan, Shay~B. Cohen, and Mirella Lapata. 2018.
\newblock \href {https://doi.org/10.18653/v1/D18-1206} {Don{'}t give me the
  details, just the summary! topic-aware convolutional neural networks for
  extreme summarization}.
\newblock In \emph{Proceedings of the 2018 Conference on Empirical Methods in
  Natural Language Processing}, pages 1797--1807, Brussels, Belgium.
  Association for Computational Linguistics.

\bibitem[{Nenkova and McKeown(2011)}]{nenkova2011automatic}
Ani Nenkova and Kathleen McKeown. 2011.
\newblock \href {https://doi.org/10.1561/1500000015} {Automatic summarization}.
\newblock \emph{Foundations and Trends® in Information Retrieval},
  5(2–3):103--233.

\bibitem[{Nenkova and McKeown(2012)}]{nenkova2012survey}
Ani Nenkova and Kathleen McKeown. 2012.
\newblock \href {https://doi.org/10.1007/978-1-4614-3223-4_3} {A survey of text
  summarization techniques}.
\newblock In \emph{Mining text data}, pages 43--76. Springer.

\bibitem[{Papineni et~al.(2002)Papineni, Roukos, Ward, and
  Zhu}]{papineni2002bleu}
Kishore Papineni, Salim Roukos, Todd Ward, and Wei-Jing Zhu. 2002.
\newblock \href {https://doi.org/10.3115/1073083.1073135} {{B}leu: a method for
  automatic evaluation of machine translation}.
\newblock In \emph{Proceedings of the 40th Annual Meeting of the Association
  for Computational Linguistics}, pages 311--318, Philadelphia, Pennsylvania,
  USA. Association for Computational Linguistics.

\bibitem[{Qi et~al.(2020)Qi, Yan, Gong, Liu, Duan, Chen, Zhang, and
  Zhou}]{yan2020prophetnet}
Weizhen Qi, Yu~Yan, Yeyun Gong, Dayiheng Liu, Nan Duan, Jiusheng Chen, Ruofei
  Zhang, and Ming Zhou. 2020.
\newblock \href {https://doi.org/10.18653/v1/2020.findings-emnlp.217}
  {{P}rophet{N}et: Predicting future n-gram for
  sequence-to-{S}equence{P}re-training}.
\newblock In \emph{Findings of the Association for Computational Linguistics:
  EMNLP 2020}, pages 2401--2410, Online. Association for Computational
  Linguistics.

\bibitem[{Raffel et~al.(2020)Raffel, Shazeer, Roberts, Lee, Narang, Matena,
  Zhou, Li, and Liu}]{raffel2019exploring}
Colin Raffel, Noam Shazeer, Adam Roberts, Katherine Lee, Sharan Narang, Michael
  Matena, Yanqi Zhou, Wei Li, and Peter~J Liu. 2020.
\newblock \href {https://jmlr.org/papers/volume21/20-074/20-074.pdf} {Exploring
  the limits of transfer learning with a unified text-to-text transformer}.
\newblock \emph{Journal of Machine Learning Research}, 21:1--67.

\bibitem[{Rothe et~al.(2020)Rothe, Narayan, and Severyn}]{rothe2020leveraging}
Sascha Rothe, Shashi Narayan, and Aliaksei Severyn. 2020.
\newblock \href {https://doi.org/10.1162/tacl_a_00313} {Leveraging pre-trained
  checkpoints for sequence generation tasks}.
\newblock \emph{Transactions of the Association for Computational Linguistics},
  8:264--280.

\bibitem[{Rush et~al.(2015)Rush, Chopra, and Weston}]{rush2015neural}
Alexander~M. Rush, Sumit Chopra, and Jason Weston. 2015.
\newblock \href {https://doi.org/10.18653/v1/D15-1044} {A neural attention
  model for abstractive sentence summarization}.
\newblock In \emph{Proceedings of the 2015 Conference on Empirical Methods in
  Natural Language Processing}, pages 379--389, Lisbon, Portugal. Association
  for Computational Linguistics.

\bibitem[{Scialom et~al.(2020)Scialom, Dray, Lamprier, Piwowarski, and
  Staiano}]{scialom2020mlsum}
Thomas Scialom, Paul-Alexis Dray, Sylvain Lamprier, Benjamin Piwowarski, and
  Jacopo Staiano. 2020.
\newblock \href {https://doi.org/10.18653/v1/2020.emnlp-main.647} {{MLSUM}: The
  multilingual summarization corpus}.
\newblock In \emph{Proceedings of the 2020 Conference on Empirical Methods in
  Natural Language Processing (EMNLP)}, pages 8051--8067, Online. Association
  for Computational Linguistics.

\bibitem[{See et~al.(2017)See, Liu, and Manning}]{see-etal-2017-get}
Abigail See, Peter~J. Liu, and Christopher~D. Manning. 2017.
\newblock \href {https://doi.org/10.18653/v1/P17-1099} {Get to the point:
  Summarization with pointer-generator networks}.
\newblock In \emph{Proceedings of the 55th Annual Meeting of the Association
  for Computational Linguistics (Volume 1: Long Papers)}, pages 1073--1083,
  Vancouver, Canada. Association for Computational Linguistics.

\bibitem[{Shazeer and Stern(2018)}]{shazeer2018adafactor}
Noam Shazeer and Mitchell Stern. 2018.
\newblock \href {http://proceedings.mlr.press/v80/shazeer18a.html} {Adafactor:
  Adaptive learning rates with sublinear memory cost}.
\newblock In \emph{Proceedings of the 35th International Conference on Machine
  Learning, {ICML} 2018}, volume~80, pages 4603--4611. {PMLR}.

\bibitem[{Sutskever et~al.(2014)Sutskever, Vinyals, and
  Le}]{sutskever2014sequence}
Ilya Sutskever, Oriol Vinyals, and Quoc~V Le. 2014.
\newblock \href
  {https://papers.nips.cc/paper/2014/file/a14ac55a4f27472c5d894ec1c3c743d2-Paper.pdf}
  {Sequence to sequence learning with neural networks}.
\newblock In \emph{Proceedings of the 27th International Conference on Neural
  Information Processing Systems (NIPS 2014)}, pages 3104--3112, Montreal,
  Canada.

\bibitem[{Tandon et~al.(2018)Tandon, Varde, and
  de~Melo}]{tandon2018commonsense}
Niket Tandon, Aparna~S Varde, and Gerard de~Melo. 2018.
\newblock \href {https://doi.org/10.1145/3186549.3186562} {Commonsense
  knowledge in machine intelligence}.
\newblock \emph{ACM SIGMOD Record}, 46(4):49--52.

\bibitem[{Vaswani et~al.(2017)Vaswani, Shazeer, Parmar, Uszkoreit, Jones,
  Gomez, Kaiser, and Polosukhin}]{vaswani2017attention}
Ashish Vaswani, Noam Shazeer, Niki Parmar, Jakob Uszkoreit, Llion Jones,
  Aidan~N Gomez, {\L}ukasz Kaiser, and Illia Polosukhin. 2017.
\newblock \href
  {https://papers.nips.cc/paper/7181-attention-is-all-you-need.pdf} {Attention
  is all you need}.
\newblock In \emph{Proceedings of the 31st International Conference on Neural
  Information Processing Systems (NIPS 2017)}, page 6000–6010, Long Beach,
  California, USA.

\bibitem[{Wang et~al.(2020)Wang, Cho, and Lewis}]{wang2020asking}
Alex Wang, Kyunghyun Cho, and Mike Lewis. 2020.
\newblock \href {https://doi.org/10.18653/v1/2020.acl-main.450} {Asking and
  answering questions to evaluate the factual consistency of summaries}.
\newblock In \emph{Proceedings of the 58th Annual Meeting of the Association
  for Computational Linguistics}, pages 5008--5020, Online. Association for
  Computational Linguistics.

\bibitem[{Widyassari et~al.(2020)Widyassari, Rustad, Shidik, Noersasongko,
  Syukur, Affandy et~al.}]{widyassari2020review}
Adhika~Pramita Widyassari, Supriadi Rustad, Guruh~Fajar Shidik, Edi
  Noersasongko, Abdul Syukur, Affandy Affandy, et~al. 2020.
\newblock \href {https://doi.org/10.1016/j.jksuci.2020.05.006} {Review of
  automatic text summarization techniques \& methods}.
\newblock \emph{Journal of King Saud University - Computer and Information
  Sciences}.

\bibitem[{Wolf et~al.(2020)Wolf, Debut, Sanh, Chaumond, Delangue, Moi, Cistac,
  Rault, Louf, Funtowicz, Davison, Shleifer, von Platen, Ma, Jernite, Plu, Xu,
  Le~Scao, Gugger, Drame, Lhoest, and Rush}]{wolf2020transformers}
Thomas Wolf, Lysandre Debut, Victor Sanh, Julien Chaumond, Clement Delangue,
  Anthony Moi, Pierric Cistac, Tim Rault, Remi Louf, Morgan Funtowicz, Joe
  Davison, Sam Shleifer, Patrick von Platen, Clara Ma, Yacine Jernite, Julien
  Plu, Canwen Xu, Teven Le~Scao, Sylvain Gugger, Mariama Drame, Quentin Lhoest,
  and Alexander Rush. 2020.
\newblock \href {https://doi.org/10.18653/v1/2020.emnlp-demos.6} {Transformers:
  State-of-the-art natural language processing}.
\newblock In \emph{Proceedings of the 2020 Conference on Empirical Methods in
  Natural Language Processing: System Demonstrations}, pages 38--45, Online.
  Association for Computational Linguistics.

\bibitem[{Wu et~al.(2016)Wu, Schuster, Chen, Le, Norouzi, Macherey, Krikun,
  Cao, Gao, Macherey et~al.}]{wu2016google}
Yonghui Wu, Mike Schuster, Zhifeng Chen, Quoc~V Le, Mohammad Norouzi, Wolfgang
  Macherey, Maxim Krikun, Yuan Cao, Qin Gao, Klaus Macherey, et~al. 2016.
\newblock \href {https://arxiv.org/abs/1609.08144} {Google's neural machine
  translation system: Bridging the gap between human and machine translation}.
\newblock \emph{arXiv:1609.08144}.

\bibitem[{Xue et~al.(2021)Xue, Constant, Roberts, Kale, Al-Rfou, Siddhant,
  Barua, and Raffel}]{xue2020mt5}
Linting Xue, Noah Constant, Adam Roberts, Mihir Kale, Rami Al-Rfou, Aditya
  Siddhant, Aditya Barua, and Colin Raffel. 2021.
\newblock \href {https://www.aclweb.org/anthology/2021.naacl-main.41} {m{T}5: A
  massively multilingual pre-trained text-to-text transformer}.
\newblock In \emph{Proceedings of the 2021 Conference of the North American
  Chapter of the Association for Computational Linguistics: Human Language
  Technologies}, pages 483--498, Online. Association for Computational
  Linguistics.

\bibitem[{Zhang et~al.(2020)Zhang, Zhao, Saleh, and Liu}]{zhang2020pegasus}
Jingqing Zhang, Yao Zhao, Mohammad Saleh, and Peter Liu. 2020.
\newblock \href {http://proceedings.mlr.press/v119/zhang20ae.html} {{PEGASUS}:
  Pre-training with extracted gap-sentences for abstractive summarization}.
\newblock In \emph{Proceedings of the 37th International Conference on Machine
  Learning, {ICML} 2020}, volume 119, pages 11328--11339. {PMLR}.

\bibitem[{Zhu et~al.(2019)Zhu, Wang, Wang, Zhou, Zhang, Wang, and
  Zong}]{zhu2019ncls}
Junnan Zhu, Qian Wang, Yining Wang, Yu~Zhou, Jiajun Zhang, Shaonan Wang, and
  Chengqing Zong. 2019.
\newblock \href {https://doi.org/10.18653/v1/D19-1302} {{NCLS}: Neural
  cross-lingual summarization}.
\newblock In \emph{Proceedings of the 2019 Conference on Empirical Methods in
  Natural Language Processing and the 9th International Joint Conference on
  Natural Language Processing (EMNLP-IJCNLP)}, pages 3054--3064, Hong Kong,
  China. Association for Computational Linguistics.

\end{thebibliography}




\end{document}